\RequirePackage{fix-cm}
\documentclass[smallextended,table]{svjour3} 
\smartqed
\usepackage[english]{babel}
\usepackage[utf8]{inputenc}
\usepackage{url}
\usepackage{tikz}
\usepackage{graphicx} 
\usepackage{amssymb}
\usepackage{multirow}
\usepackage{adjustbox}
\usepackage{booktabs}
\usepackage{tabularx}
\usepackage{natbib}
\usetikzlibrary{quotes,arrows.meta,positioning,decorations.pathreplacing,calc,3d,arrows}

\usepackage{subfig}
\usepackage[misc]{ifsym}

\usepackage[keeplastbox]{flushend}
\usepackage[a4paper=false,
            citecolor=blue,
            colorlinks=true,
            urlcolor=blue,
            linkcolor=blue,
            pdfauthor={Hassan Ismail Fawaz},
            pdftitle={InceptionTime: Finding AlexNet for Time Series Classification},
            pdfsubject={InceptionTime: Finding AlexNet for Time Series Classification},
            pdfkeywords={deep learning, time series, classification}
            ]{hyperref}
\usepackage{textcomp}
\usepackage{tikz}
\newcommand\copyrighttext{%
\small This is the author's version of an article published in Data Mining and Knowledge Discovery.
The final authenticated version is available online at: \url{https://doi.org/10.1007/s10618-020-00710-y}.}
\newcommand\copyrightnotice{%
\begin{tikzpicture}[remember picture,overlay]
\node[anchor=north,yshift=-30pt] at (current page.north) {\fbox{\parbox{\textwidth}{\copyrighttext}}};
\end{tikzpicture}
}

\usepackage[11pt]{extsizes}

\usepackage{geometry}
\newcommand{\subparagraph}{}
\usepackage{titlesec}
\titleformat*{\section}{\sffamily\Large\bfseries}
\titleformat*{\subsection}{\sffamily\large\bfseries}
\titleformat*{\subsubsection}{\sffamily\large\bfseries}

\usepackage{enumitem}
\usepackage{setspace}
\setlist{itemsep=0pt,parsep=0pt}

\begin{document}

\hyphenation{avai-lability}

\def\makeheadbox{\relax}
\newcommand{\gfc}[1]{{\color{magenta}~{\bf [Germain:} #1{\bf ]}}}
\newcommand{\hfc}[1]{{\color{blue}~{\bf [Hassan:} #1{\bf ]}}}
\newcommand{\jwc}[1]{{\color{teal}~{\bf [Jonathan:} #1{\bf ]}}}
\newcommand{\bl}[1]{{\color{red}~{\bf [Ben:} #1{\bf ]}}}
\newcommand{\cpc}[1]{{\color{cyan}~{\bf [Charlotte:} #1{\bf ]}}}
\newcommand{\fpc}[1]{{\color{orange}~{\bf [Francois:} #1{\bf ]}}}
\newcommand{\gw}[1]{{\color{purple}~{\bf [Geoff:} #1{\bf ]}}}
\newcommand{\dfs}[1]{{\color{pink}~{\bf [Daniel:} #1{\bf ]}}}
\newcommand{\ourmethod}[1]{InceptionTime}

\newcommand{\revisionn}[1]{{\color{black}#1}}

\title{\ourmethod{}: Finding AlexNet for Time Series Classification}
\titlerunning{\ourmethod{}: Finding AlexNet for Time Series Classification}

\author{Hassan Ismail Fawaz\textsuperscript{1} \and
Benjamin Lucas\textsuperscript{2} \and 
Germain Forestier\textsuperscript{1,2} \and
Charlotte Pelletier\textsuperscript{2,3} \and 
Daniel F. Schmidt\textsuperscript{2} \and 
Jonathan Weber\textsuperscript{1} \and 
Geoffrey I. Webb\textsuperscript{2} \and 
Lhassane Idoumghar\textsuperscript{1} \and 
Pierre-Alain Muller\textsuperscript{1} \and 
Fran\c{c}ois Petitjean\textsuperscript{2}
}

\authorrunning{Hassan Ismail Fawaz et al.} 
\institute{\Letter{} H. Ismail Fawaz  \\ \medskip
              \email{hassan.ismail-fawaz@uha.fr}  \\
              \textsuperscript{1}IRIMAS, Universit\'e Haute Alsace, Mulhouse, France\\\
              \textsuperscript{2}Faculty of IT, Monash University, Melbourne, Australia \\
                            \textsuperscript{3}IRISA, UMR CNRS 6074, Universit\'e Bretagne Sud, Vannes, France
}

\date{}
\sloppy
\maketitle    

\copyrightnotice{}              

\begin{abstract}
This paper brings deep learning at the forefront of research into Time Series Classification (TSC). 
TSC is the area of machine learning tasked with the categorization (or labelling) of time series. 
The last few decades of work in this area have led to significant progress in the accuracy of classifiers, with the state of the art now represented by the HIVE-COTE algorithm. 
While extremely accurate, HIVE-COTE cannot be applied to many real-world datasets because of its high training time complexity in $O(N^2\cdot T^4)$ for a dataset with $N$ time series of length $T$. 
For example, it takes HIVE-COTE more than 8 days to learn from a small dataset with $N=1500$ time series of short length $T=46$. 
Meanwhile deep learning has received enormous attention because of its high accuracy and scalability. 
\revisionn{Recent approaches to deep learning for TSC have been scalable, but less accurate than HIVE-COTE.}
We introduce \ourmethod{} --- an ensemble of deep Convolutional Neural Network (CNN) models, inspired by the Inception-v4 architecture.
Our experiments show that \ourmethod{} is on par with HIVE-COTE in terms of accuracy while being much more scalable: not only can it learn from 1,500 time series in one hour but it can also learn from 8M time series in 13 hours, a quantity of data that is fully out of reach of HIVE-COTE.
\keywords{time series classification \and deep learning \and scalable model \and inception}
\end{abstract}

\section{Introduction}

Recent times have seen an explosion in the magnitude and prevalence of time series data.
Industries varying from health care~\citep{forestier2018surgical,lee2018diagnosis,IsmailFawaz2019automatic} and social security~\citep{yi2018an} to human activity recognition~\citep{yuan2018muvan} and remote sensing~\citep{pelletier2019temporal}, all now produce time series datasets of previously unseen scale --- both in terms of time series length and quantity.
This growth also means an increased dependence on automatic classification of time series data, and ideally, algorithms with the ability to do this at scale.

These problems, known as Time Series Classification (TSC), differ significantly to traditional supervised learning for structured data, in that the algorithms should be able to handle and harness the temporal information present in the signal~\citep{bagnall2017the}.
It is easy to draw parallels from this scenario to computer vision problems such as image classification and object localization, where successful algorithms learn from the spatial information contained in an image.
Put simply, the time series problem is essentially the same class of problem, just with one less dimension.
Yet despite this similarity, the current state-of-the-art algorithms from the two fields share little resemblance~\citep{ismailfawaz2018deep}.

Deep learning has a long history (in machine learning terms) in computer vision~\citep{lecun1998efficient} but its popularity exploded with AlexNet~\citep{krizhevsky2012imagenet}, after which it has been unquestionably the most successful class of algorithms~\citep{lecun2015deep}. 
Conversely, deep learning has only recently started to gain popularity amongst time series data mining researchers~\citep{ismailfawaz2018deep}.
This is emphasized by the fact that the Residual Network (ResNet), which is currently considered the state-of-the-art neural network architecture for TSC when evaluated on the UCR archive~\citep{dau2018ucr}, was originally proposed merely as a baseline model for the underlying task~\citep{wang2017time}.
Given the similarities in the data, it is easy to suggest that there is much potential improvement for deep learning in TSC.

In this paper, we take an important step towards finding the equivalent of `AlexNet' for TSC by presenting \ourmethod{} --- a novel deep learning ensemble for TSC.
\ourmethod{} achieves state-of-the-art accuracy when evaluated on the UCR archive (currently the largest publicly available repository for TSC~\citep{dau2018ucr}) while also possessing ability to scale to a magnitude far beyond that of its strongest competitor.

\ourmethod{} is an ensemble of five deep learning models for TSC, each one created by cascading multiple Inception modules~\citep{szegedy2015going}. 
Each individual classifier (model) will have exactly the same architecture but with different randomly initialized weight values.
The core idea of an Inception module is to apply multiple filters simultaneously to an input time series.
The module includes filters of varying lengths, which as we will show, allows the network to automatically extract relevant features from both long and short time series.

After presenting \ourmethod{} and its results, we perform an analysis of the architectural hyperparameters of deep neural networks --- depth, filter length, number of filters --- and the characteristics of the Inception module --- the bottleneck and residual connection, in order to provide insight into why this model is so successful.
In fact, we construct networks with filters larger than have ever been explored for computer vision tasks, taking direct advantage of the fact that time series exhibit one less dimension than images.


The remainder of this paper is structured as follows: first we start by presenting the background and related work in Section~\ref{sec:background}.
We then proceed in Section~\ref{sec:method} to explain the network architecture and its main building block --- the Inception module.
Section~\ref{sec:exp} contains the details of our experimental setup. 
In Section~\ref{sec:experiments}, we show that \ourmethod{} produces state-of-the-art accuracy on the UCR archive, the TSC benchmark, while also presenting a runtime comparison with its nearest competitor.
In Section~\ref{sec:hyp}, we provide a detailed hyperparameter study that provides insight into the choices made when designing our proposed neural network. Finally we conclude the paper in Section~\ref{sec:conclusion} and give directions for further research on deep learning for TSC.

\section{Related work}\label{sec:background}
In this section, we start with some preliminary definitions for ease of understanding, before presenting the current state-of-the-art algorithms for TSC. 
We end by providing a deeper background for designing neural network architectures for domain-agnostic TSC problems.

\subsection{Time series classification}

\begin{definition}
A Multivariate Time Series (MTS) $X=[X_1,X_2, \dots, X_T]$ with $M$ dimensions, consists of $T$ ordered elements $X_i$ $\in \mathbb{R}^M$.
\end{definition}

\begin{definition}
A \emph{Univariate} time series $X$ of length $T$ is simply an MTS with $M=1$. 
\end{definition}

\begin{definition}
$D=\{(X^1,Y^1), (X^2,Y^2), \dots, (X^N,Y^N)\}$ is a dataset containing a collection of pairs $(X^i,Y^i)$ where $X^i$ could either be a univariate or multivariate time series with its corresponding label denoted by $Y^i$.
\end{definition}

The task of classifying time series data consists of learning a classifier on $D$ in order to map from the space of possible inputs $X$ to a probability distribution over the classes $Y$.

The state-of-the-art for TSC has been organized by~\cite{bagnall2017the} into the following main categories:

\subsubsection{Whole series}
This type of classifiers compares two series using a certain distance.
For many years, the leading classifier for TSC was the nearest neighbor algorithm coupled with the Dynamic Time Warping similarity measure (NN-DTW)~\citep{bagnall2017the}.
Much research has subsequently focused on finding alternative similarity measures~\citep{marteau2009time, stefan2013the, keogh2001derivative, vlachos2006indexing}, however none have been found to significantly outperform NN-DTW on the UCR Archive~\citep{lines2015time}.
Another research area focused on proposing global alignment kernels such as SoftDTW introduced by~\cite{cuturi2017soft}, that can be further used in a nearest centroid classification scheme.
This research informed one current state-of-the-art method, named Elastic Ensemble (EE), which is an ensemble of 11 nearest neighbor classifiers each coupled with a different similarity measure~\citep{lines2015time}.
While this algorithm produces state-of-the-art accuracy, its use on large datasets is limited by its training complexity, with some of its parameter searches being in $O(N^2\cdot T^3)$.
Following this line of research, all recent successful classification algorithms for time series data are all ensemble based models.
Furthermore, to tackle EE's huge training time,~\cite{lucas2019proximity} proposed a tree-based ensemble called Proximity Forest (PF) that uses EE's distances as a splitting criteria while replacing the parameter searches by a random sampling.

\subsubsection{Dictionary based}
This type of classifiers discriminate time series by the frequency of repetition of some sub-series. 
The most famous one being the Bag-of-SFA-Symbols (BOSS), which is based on an ensemble of NNs classifiers coupled with a bespoke Euclidean distance computed on the frequency histograms obtained from the Symbolic Fourier Approximation (SFA) discretization~\citep{schafer2015the}.
BOSS has a high training complexity of $O(N^2)$, which the authors identified as a shortcoming and attempted to address with subsequent scalable variations of the algorithm in~\cite{schafer2015scalable, Schafer2017WEASEL}, however neither of these reached state-of-the-art accuracy.

\subsubsection{Shapelets}
This family of algorithms focuses on finding relatively short repeated subsequences to identify a certain class. 
These patterns are time independent and are called shapelets.
Another type of ensemble classifiers is shapelet based algorithms, such as in~\cite{hills2014classification}, where discriminative subsequences (shapelets) are extracted from the training set and fed to off-the-shelf classifiers such as Support Vector Machines and Random Forests.
The shapelet transform has a training complexity of $O(N^2\cdot T^4)$ and thus, again, has little potential to scale to large datasets.

\subsubsection{Transformation ensembles}
More recently,~\cite{bagnall2016time} noted that there is no single time series transformation technique (such as shapelets or SFA) that significantly dominates the others, showing that constructing an ensemble of different classifiers over different time series representations, called COTE, will significantly improve the accuracy.
\cite{lines2016hive} extended COTE with a hierarchical voting scheme, which further improves the decision taken by the ensemble. 
Named the Hierarchical Vote Collective of Transformation-Based Ensembles (HIVE-COTE), it represents the current state-of-the-art accuracy when evaluated on the UCR archive, however its practicality is hindered by its huge training complexity of order $O(N^2 \cdot T^4)$.
This is highlighted by the extensive experiments in~\cite{lucas2019proximity} where PF showed competitive performance with COTE, while having a runtime that is orders of magnitudes lower.
Deep learning models, which we will discuss in detail in the following subsection, also significantly beat the runtime of HIVE-COTE by trivially leveraging GPU parallel computation abilities.
A~comprehensive detailed review of recent methods for TSC can be found in~\cite{bagnall2017the}. 

\subsection{Deep learning for time series classification}
Since the recent success of deep learning techniques in supervised learning such as image recognition~\citep{zhang2018similarity} and natural language processing~\citep{guan2019towards}, researchers started investigating these complex machine learning models for TSC~\citep{wang2017time,cui2016multi,IsmailFawaz2019adversarial}.
Precisely, Convolutional Neural Networks (CNNs) have showed promising results for TSC. 
Given an input MTS, a convolutional layer consists of sliding one-dimensional filters over the time series, thus enabling the network to extract non-linear discriminant features that are time-invariant and useful for classification.
By cascading multiple layers, the network is able to further extract hierarchical features that should in theory improve the network's prediction. 
Note that given an input univariate time series, by applying several one-dimensional filters, the outcome can be considered an MTS whose length is preserved and the number of dimensions $M$ is equal the number of filters applied at this layer. 
More details on how deep CNNs are being adapted for one-dimensional time series data can be found in~\cite{ismailfawaz2018deep}. 
The rest of this subsection is dedicated to describing what is currently being explored in deep learning for TSC.

Multi-scale Convolutional Neural Networks (MCNN)~\citep{cui2016multi} and Time LeNet~\citep{leguennec2016data} are considered among the first architectures to be validated on a domain-agnostic TSC benchmark such as the UCR archive.
These models were inspired by image recognition modules, which hindered their accuracy, mainly because of the use of progressive pooling layers, that were mainly added for computational feasibility when dealing with image data~\citep{sabour2017dynamic}.
Consequently, Fully Convolutional Neural Networks (FCNs) were shown to achieve great performance without the need to add pooling layers to reduce the input data's dimensionality~\citep{wang2017time}. 
More recently, it has been shown that deeper CNN models coupled with residual connections such as ResNet can further improve the classification performance~\citep{ismailfawaz2018deep}. 
In essence, since time series data exhibit only one structuring dimension (i.e. time, as opposed to two spatial dimensions for images), it is possible to explore more complex models that are usually computationally infeasible for image recognition problems: for example removing the pooling layers that throw away valuable information in favour of reducing the model's complexity. 
In this paper, we propose an Inception based network that applies several convolutions with various filters lengths. In contrast to networks designed for images, we are able to explore filters 10 times longer than recent Inception variants for image recognition tasks~\citep{szegedy2017inception}.  

Inception was first proposed by~\cite{szegedy2015going} for end-to-end image classification. 
Now the network has evolved to become Inceptionv4, where Inception was coupled with residual connections to further improve the performance~\citep{szegedy2017inception}. 
As for TSC a relatively competitive Inception-based approach was proposed in~\cite{karimi2018scalable}, where time series where transformed to images using Gramian Angular Difference Field (GADF), and finally fed to an Inception model that had been pre-trained for (standard) image recognition. 
Unlike this feature engineering approach, by adopting an end-to-end learning from raw time series data, a one-dimensional Inception model was used for Supernovae classification using the light flux of a region in space as an input MTS for the network~\citep{brunel2019a}.
However, the authors limited the conception of their Inception architecture to the one proposed by Google for ImageNet~\citep{szegedy2017inception}. 
In our work, we explore much larger filters than any previously proposed network for TSC in order to reach state-of-the-art performance on the UCR benchmark.  


\section{InceptionTime: an accurate and scalable time series classifier}\label{sec:method}
In this section, we start by describing the proposed architecture we call \ourmethod{} for classifying time series data. Specifically, we detail the main component of our network: the Inception module. 
We then present our proposed model \ourmethod{} which consists of an ensemble of 5 different Inception networks initialized randomly.
Finally, we adapt the concept of Receptive Field for time series data.

\begin{figure}
    \centering
    \includegraphics[width=0.9\linewidth]{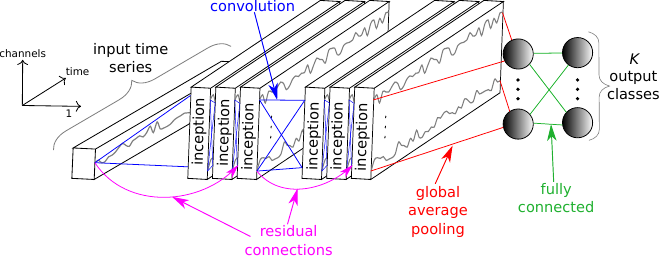}
    \caption{Our Inception network for time series classification}
    \label{fig:inception-archi}
\end{figure}

\subsection{Inception Network: a novel architecture for TSC}
The composition of an Inception network classifier contains \emph{two} different residual blocks, as opposed to ResNet, which is comprised of \emph{three}.
For the Inception network, each block is comprised of three Inception modules rather than traditional fully convolutional layers.
Each residual block's input is transferred via a shortcut linear connection to be added to the next block's input, thus mitigating the vanishing gradient problem by allowing a direct flow of the gradient~\citep{he2016deep}. 
Following these residual blocks, we employed a Global Average Pooling (GAP) layer that averages the output multivariate time series over the whole time dimension.
\revisionn{At last, we used a final traditional fully-connected softmax layer with a number of neurons equal to the number of classes in the dataset.}
\figurename~\ref{fig:inception-archi} depicts an Inception network's architecture showing 6 different Inception modules stacked one after the other.

\begin{figure*}
    \centering
    \includegraphics[width=0.85\linewidth]{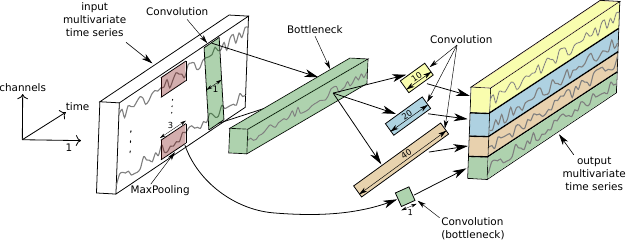}
    \caption{Inside our Inception module for time series classification. 
    For simplicity we illustrate a bottleneck layer of size $m=1$.}
    \label{fig:inception-module}
\end{figure*}

As for the Inception module, \figurename~\ref{fig:inception-module} illustrates the inside details of this operation. 
\revisionn{Let us consider the input to be an MTS with $M$ dimensions.} 
The first major component of the Inception module is called the ``bottleneck'' layer.
This layer performs an operation of sliding $m$ filters of length 1 with a stride equal to 1.
This will transform the time series from an MTS with $M$ dimensions to an MTS with $m \ll M$ dimensions, thus reducing significantly the dimensionality of the time series as well as the model's complexity and mitigating overfitting problems for small datasets.
Note that for visualization purposes, \figurename~\ref{fig:inception-module} illustrates a bottleneck layer with $m=1$. 
Finally, we should mention that this bottleneck technique allows the Inception network to have much longer filters than ResNet (almost ten times) with roughly the same number of parameters to be learned, since without the bottleneck layer, the filters will have $M$ dimensions compared to $m \ll M$ when using the bottleneck layer. 
The second major component of the Inception module is sliding multiple filters of different lengths simultaneously on the same input time series. 
For example in \figurename~\ref{fig:inception-module}, three different convolutions with length $l \in \{10,20,40\}$ are applied to the input MTS, which is technically the output of the bottleneck layer. 
Additionally, in order to make our model invariant to small perturbations, we introduce another parallel MaxPooling operation, followed by a bottleneck layer to reduce the dimensionality.
The output of sliding a MaxPooling window is computed by taking the maximum value in this given window of time series. 
Finally, the output of each independent parallel convolution/MaxPooling is concatenated to form the output MTS.
The latter operations are repeated for each individual Inception module of the proposed network.

By stacking multiple Inception modules and training the weights (filters' values) via backpropagation, the network is able to extract latent hierarchical features of multiple resolutions thanks to the use of filters with various lengths.
For completeness, we specify the exact number of filters for our proposed Inception module: 3 sets of filters each with 32 filters of length $l\in \{10,20,40\}$ with MaxPooling added to the mix, thus making the total number of filters per layer equal to $32\times4=128=M$ - the dimensionality of the output MTS. 
The default bottleneck size value was set to $m=32$.

\subsection{\ourmethod{}: a neural network ensemble for TSC}
Our proposed state-of-the-art \ourmethod{} model is an ensemble of 5 Inception networks, with each prediction given an even weight. 
In fact, during our experimentation, we have noticed that a single Inception network exhibits high standard deviation in accuracy, which is very similar to ResNet's behavior~\citep{IsmailFawaz2019deep}.
We believe that this variability comes from both the randomly initialized weights and the stochastic optimization process itself. 
This was an important finding for us, previously observed in~\cite{scardapane2017randomness}, as rather than training only one, potentially very good or very poor, instance of the Inception network, we decided to leverage this instability through ensembling, creating \ourmethod{}.
The following equation explains the ensembling of predictions made by a network with different initializations:
\begin{equation}
    \hat{y}_{i,c}=\frac{1}{n}\sum_{j=1}^{n}\sigma_c(x_i,\theta_j) ~~|~~\forall c\in [1,C]
\end{equation}
with $\hat{y}_{i,c}$ denoting the ensemble's output probability of having the input time series $x_i$ belonging to class $c$, which is equal to the logistic output $\sigma_c$ averaged over the $n$ randomly initialized models.
More details on ensembling neural networks for TSC can be found in~\cite{IsmailFawaz2019deep}. 
As for the proposed model in this paper, we chose the number of individual classifiers to be equal to $5$, which is justified in Section~\ref{sec:experiments}.
We should note that we have opted to a neural network ensemble given the small training size of the UCR archive datasets which are not well suited to deep learning approaches, thus allowing us to control and leverage the variance of the error, which is likely to reduce when increasing the training set's size.

\subsection{Receptive field}

\begin{figure}
    \centering
    \includegraphics[width=0.7\linewidth]{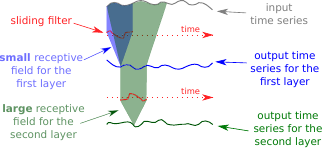}
    \caption{Receptive field illustration for a two layers CNN}
    \label{fig:receptive-field}
\end{figure}

The concept of Receptive Field (RF) is an essential tool to the understanding of deep CNNs~\citep{luo2016understanding}.
Unlike fully-connected networks or Multi-Layer Perceptrons, a neuron in a CNN depends only on a region of the input signal. 
This region in the input space is called the receptive field of that particular neuron. 
For computer vision problems this concept was extensively studied, such as in~\cite{liu2018understanding} where the authors compared the effective and theoretical receptive fields of a CNN for image segmentation.   

For temporal data, the receptive field can be considered as a theoretical value that measures the maximum field of view of a neural network in a one-dimensional space: the larger it is, the better the network becomes (in theory) in detecting longer patterns. 
We now provide the definition of the RF for time series data, which is later used in our experiments. 
Suppose that we are sliding convolutions with a stride equal to $1$.
The formula to compute the RF for a network of depth $d$ with each layer having a filter length equal to $k_i$ with $i\in[1,d]$ is: 
\begin{equation}\label{eq-rf}
    1+\sum_{i=1}^{d}(k_i-1)
\end{equation}
By analyzing equation~\ref{eq-rf} we can clearly see that adding two layers to the initial set of $d$ layers, will increase only slightly the value of $RF$. 
In fact in this case, if the old $RF$ value is equal to $RF^{'}$, the new value $RF$ will be equal to $RF^{'} + 2\times (k-1)$. 
Conversely, by increasing the filter length $k_i$, $\forall i \in [1,d]$ by 2, the new value $RF$ will be equal to $RF^{'} + 2\times d$.
This is rather expected since by increasing the filter length for all layers, we are actually increasing the $RF$ for each layer in the network.
\figurename~\ref{fig:receptive-field} illustrates the RF for a two layers CNN. 

In this paper, we chose to focus on the RF concept since it has been known for computer vision problems, that larger RFs are required to capture more context for object recognition~\citep{luo2016understanding}.
Following the same line of thinking, we hypothesize that detecting larger patterns from very long one-dimensional time series data, requires larger receptive fields. 

\begin{figure}
    \centering
    \includegraphics[width=0.5\linewidth]{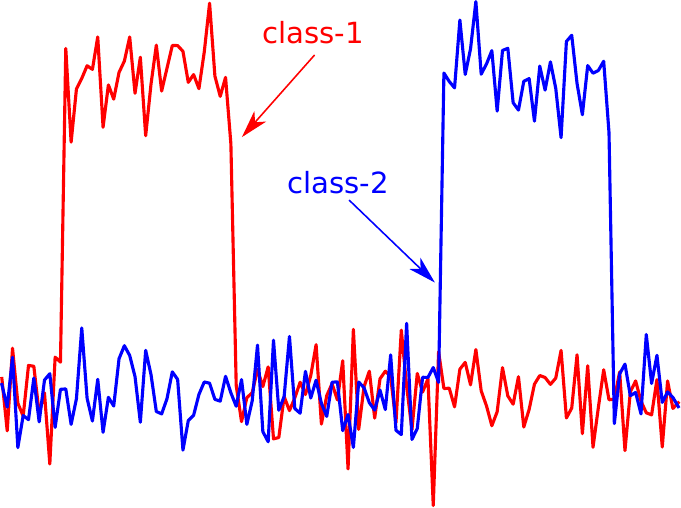}
    \caption{Example of a synthetic binary time series classification problem}
    \label{fig:synthetic-dataset-example}
\end{figure}

\section{Experimental setup}\label{sec:exp}

First, we detail the method to generate our synthetic dataset, which is later used in our architecture and hyperparameter study. 
For testing our different deep learning methods, we created our own synthetic TSC dataset. 
The goal was to be able to control the length of the time series data as well as the number of classes and their distribution in time.
To this end, we start by generating a univariate time series using uniformly distributed noise sampled between 0.0 and 0.1. 
Then in order to assign this synthetic random time series to a certain class, we inject a pattern with an amplitude equal to 1.0 in a pre-defined region of the time series. 
This region will be specific to a certain class, therefore by changing the placement of this pattern we can generate an unlimited amount of classes, whereas the random noise will allow us to generate an unlimited amount of time series instances per class.
One final note is that we have fixed the length of the pattern to be equal to 10\% the length of the synthetic time series. 
An example of a synthetic binary TSC problem is depicted in \figurename~\ref{fig:synthetic-dataset-example}.

All deep neural networks were trained by leveraging the parallel computation of a remote cluster of more than 60 GPUs comprised of GTX 1080 Ti, Tesla K20, K40 and K80.
Local testing and development was performed on an NVIDIA Quadro P6000. 
The latter graphics card was also used for computing the training time of a model. 
When evaluating global accuracy and computational complexity, we have used the UCR archive~\citep{dau2018ucr}, which is the largest publicly available archive for TSC.
The models were trained/tested using the original training/testing splits provided in the archive.  
To study the effect of different hyperparameters and architectural designs, we used in addition to the traditional UCR benchmark for TSC, the synthetic dataset whose generation is described in details in the previous paragraph. 
All time series data were $z$-normalized (including the synthetic series) to have a mean equal to zero and a standard deviation equal to one. 
This is considered a common best-practice before classifying time series data~\citep{bagnall2017the}. 
Finally, we should note that all models are trained using the Adam optimization algorithm~\citep{kingma2015adam} and all weights are initialized randomly using Glorot's uniform technique~\citep{glorot2010understanding}.

Similarly to~\cite{ismailfawaz2018deep}, when comparing with the state-of-the-art results published in~\cite{bagnall2017the} we used the deep learning model's median test accuracy over the different runs.
Following the recommendations in~\cite{demsar2006statistical} we adopted the Friedman test~\citep{friedman1940a} in order to reject the null hypothesis. 
We then performed the pairwise post-hoc analysis recommended by~\cite{benavoli2016should} where we replaced the average rank comparison by a Wilcoxon signed-rank test with Holm's alpha ($5\%$) correction~\citep{garcia2008an}. 
To visualize this type of comparison we used a critical difference diagram proposed by~\cite{demsar2006statistical}, where a thick horizontal line shows a cluster of classifiers (a clique) that are not-significantly different in terms of accuracy.

In order to allow for the time series community to build upon and verify our findings, the source code for all these experiments was made publicly available on our companion repository\footnote{\url{https://github.com/hfawaz/\ourmethod{}}}.
In addition, we provide the pre-trained deep learning models, thus allowing data mining practitioners to leverage these models in a transfer learning setting~\citep{IsmailFawaz2018transfer}. 

\section{Experiments: \ourmethod{}}~\label{sec:experiments}
\begin{figure}
    \centering
    \includegraphics[width=.7\linewidth]{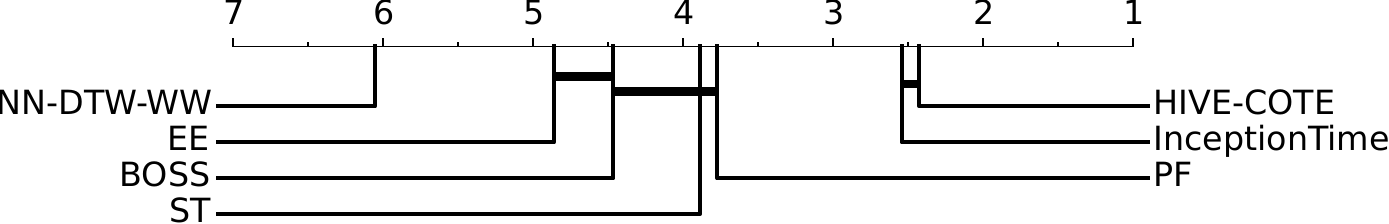}
    \caption{Critical difference diagram showing the performance of \ourmethod{} compared to the current state-of-the-art classifiers of time series data.}
    \label{fig:cd-diagram-with-inception-ensemble}
\end{figure}
In this section, we present the results of our proposed novel classifier called \ourmethod{}, evaluated on the 85 datasets of the UCR archive.
We note that throughout the paper (unless specified otherwise) \ourmethod{} refers to an ensemble of 5 Inception networks, while the ``\ourmethod{}($n$)'' notation is used to denote an ensemble of $n$ Inception networks. 

\figurename~\ref{fig:cd-diagram-with-inception-ensemble} illustrates the critical difference diagram with \ourmethod{} added to the mix of the current state-of-the-art classifiers for time series data, whose results were taken from~\cite{bagnall2017the}. 
We can see here that our \ourmethod{} ensemble reaches competitive accuracy with the class-leading algorithm HIVE-COTE, an ensemble of 37 TSC algorithms with a hierarchical voting scheme~\citep{lines2016hive}. 
While the two algorithms share the same clique on the critical difference diagram, the trivial GPU parallelization of deep learning models makes learning our \ourmethod{} model a substantially easier task than training the 37 different classifiers of HIVE-COTE, whose implementation does not trivially leverage the GPUs' computational power. 

\begin{figure}
    \centering
    \includegraphics[width=0.6\linewidth]{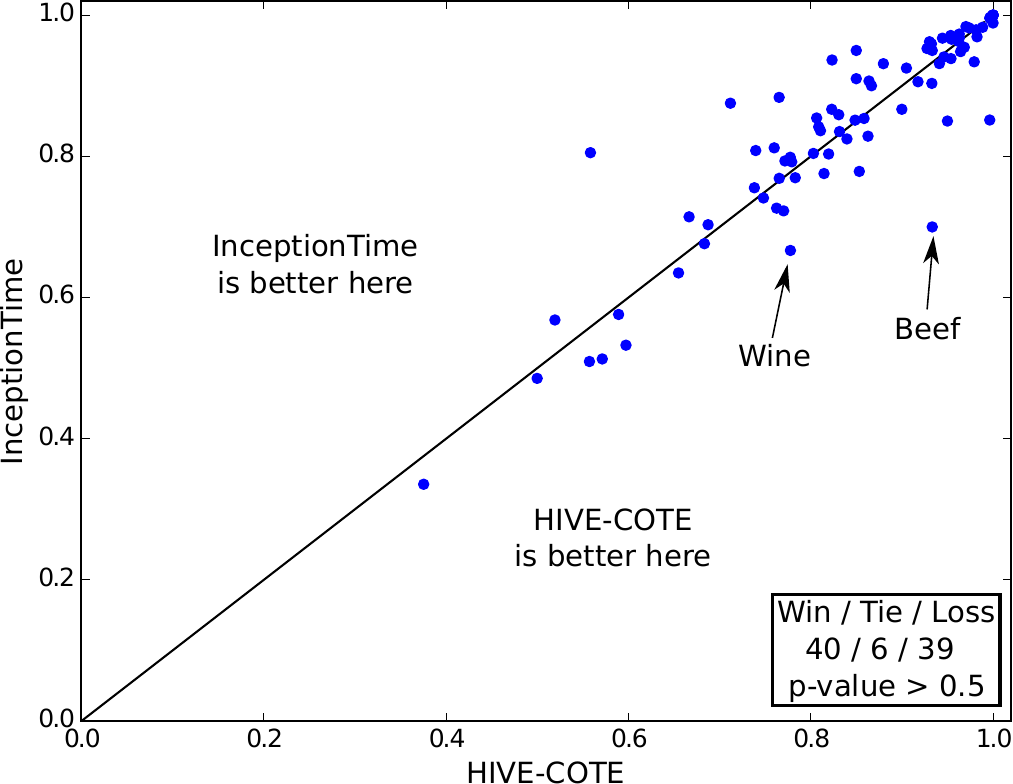}
    \caption{Accuracy plot showing how our proposed \ourmethod{} model is not significantly different than HIVE-COTE.}
    \label{fig:plot-inceptions-vs-hive-cote}
\end{figure}

To further visualize the difference between the \ourmethod{} and HIVE-COTE, \figurename~\ref{fig:plot-inceptions-vs-hive-cote} depicts the accuracy plot of \ourmethod{} against HIVE-COTE for each of the 85 UCR datasets. 
The results show a Win/Tie/Loss of 40/6/39 in favor of \ourmethod{}, however the difference is not statistically significant as previously discussed.
From \figurename~\ref{fig:plot-inceptions-vs-hive-cote}, we can also easily spot the two datasets for which  \ourmethod{} noticeably under-performs (in terms of accuracy) with respect to HIVE-COTE: Wine and Beef.
These two datasets contain spectrography data from different types of beef/wine, with the goal being to determine the correct type of meat/wine using the recorded time series data.
Recently, transfer learning has been shown to significantly increase the accuracy for these two datasets, especially when fine-tuning a dataset with similar time series data~\citep{IsmailFawaz2018transfer}. 
Our results suggest that further potential improvements may be available for \ourmethod{} when applying a transfer learning approach, as recent discoveries in~\cite{kashiparekh2019convtimenet} show that the various filter lengths of the Inception modules have been shown to benefit more from fine-tuning than networks with a static filter length.

\begin{figure}
    \centering
    \includegraphics[width=0.6\linewidth]{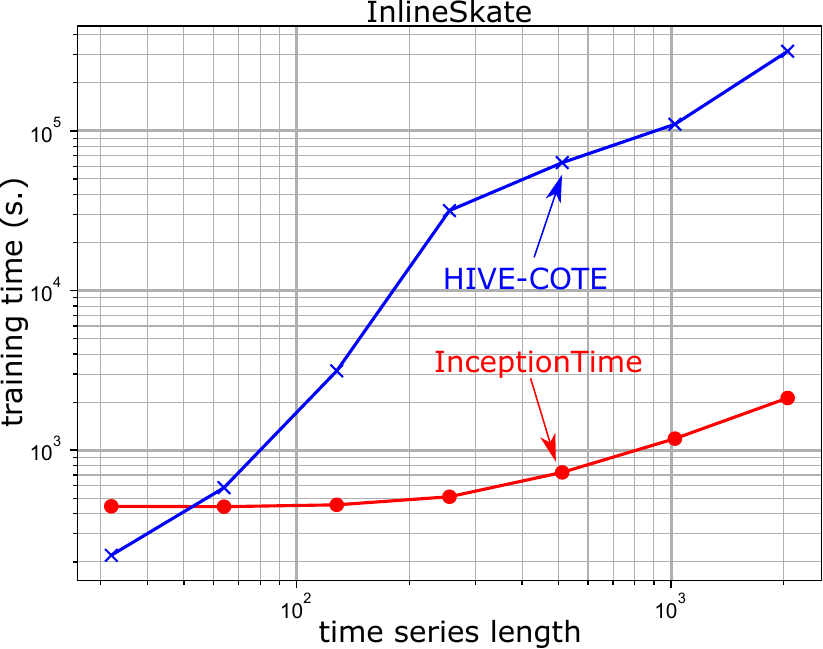}
    \caption{Training time as a function of the series length for the InlineSkate dataset.}
    \label{fig:plot-inception-vs-hive-cote-length}
\end{figure}

\begin{figure}
    \centering
    \includegraphics[width=0.6\linewidth]{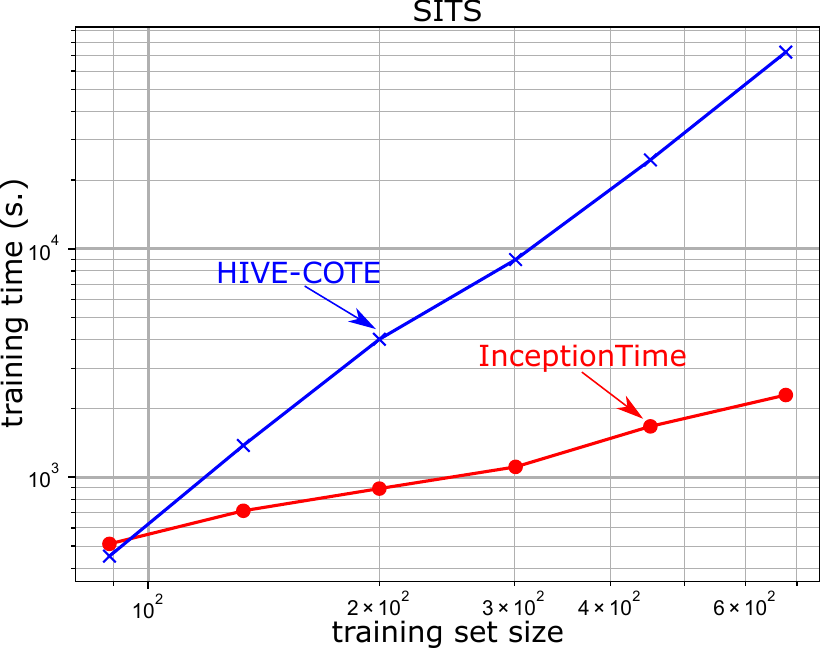}
    \caption{Training time as a function of the training set size for the SITS dataset.}
    \label{fig:plot-inception-vs-hive-cote-train-size}
\end{figure}

Now that we have demonstrated that our proposed technique is able to reach the current state-of-the-art accuracy for TSC problems, we will further investigate the time complexity of our model. 
Note that during the following experiments, we ran our ensemble on a single Nvidia Quadro P6000 in a sequential manner, meaning that for \ourmethod{}, 5 different Inception networks were trained one after the other. 
Therefore we did not make use of our remote cluster of GPUs.
First we start by investigating how our algorithm scales with respect to the length of the input time series. 
\figurename~\ref{fig:plot-inception-vs-hive-cote-length} shows the training time versus the length of the input time series. 
For this experiment, we used the InlineSkate dataset with an exponential re-sampling. 
We can clearly see that \ourmethod{}'s complexity increases almost linearly with an increase in the time series' length, unlike HIVE-COTE, whose execution is almost two order of magnitudes slower. 
Having showed that \ourmethod{} is significantly faster when dealing with long time series, we now proceed to evaluating the training time with respect to the number of time series in a dataset.
To this end, we used a Satellite Image Time Series dataset~\citep{tan2017indexing}. 
The data contain approximately one million time series, each of length 46 and labelled as one of 24 possible land-use classes (e.g.\ `wheat', `corn', `plantation', `urban').
From \figurename~\ref{fig:plot-inception-vs-hive-cote-train-size} we can easily see how our \ourmethod{} is an order of magnitude faster than HIVE-COTE, and the trend suggests that this difference will only continue to grow, rendering \ourmethod{} a clear favorite classifier in the Big Data era.  
\revisionn{
Note that HIVE-COTE uses heuristics in its implementation, which explains why the complexity appears lower in the experiments than the expected $O(T^4)$.
To summarize, we believe that \ourmethod{} should be considered as one of the top state-of-the-art methods for TSC, given that it demonstrates equal accuracy to that of HIVE-COTE (see \figurename~\ref{fig:plot-inceptions-vs-hive-cote}) while being much faster (see \figurename~\ref{fig:plot-inception-vs-hive-cote-length} and~\ref{fig:plot-inception-vs-hive-cote-train-size}). 
}

\begin{figure}
    \centering
    \includegraphics[width=0.6\linewidth]{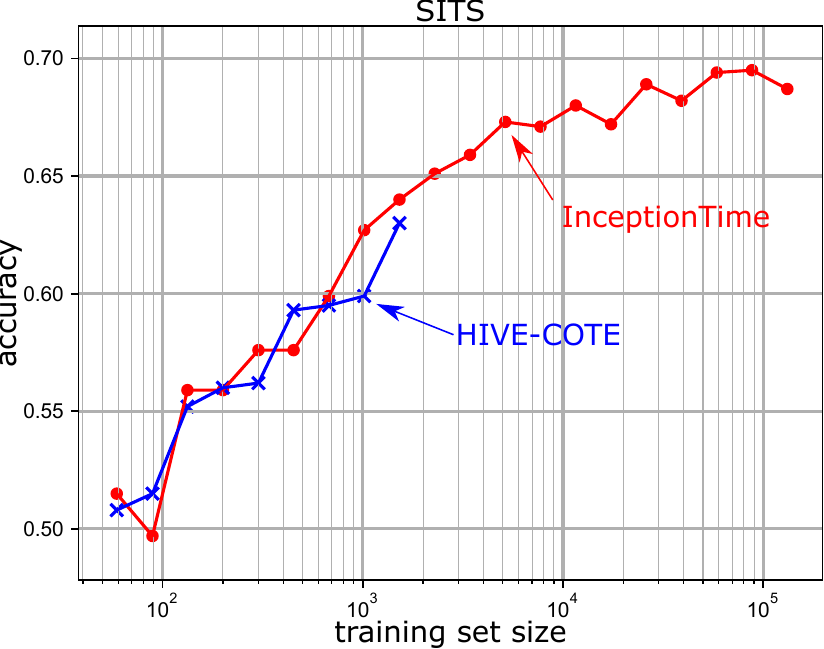}
    \caption{Accuracy as a function of the training set size for the SITS dataset.}
    \label{fig:plot-inception-vs-hive-cote-accuracy}
\end{figure}

In order to further demonstrate the capability of \ourmethod{} to handle efficiently a large amount of training samples unlike its counterpart HIVE-COTE, we show in \figurename~\ref{fig:plot-inception-vs-hive-cote-accuracy} how the accuracy continues to increase with \ourmethod{} for larger training set sizes, where HIVE-COTE would take 100 times longer to run. 

\begin{figure}
    \centering
    \includegraphics[width=0.6\linewidth]{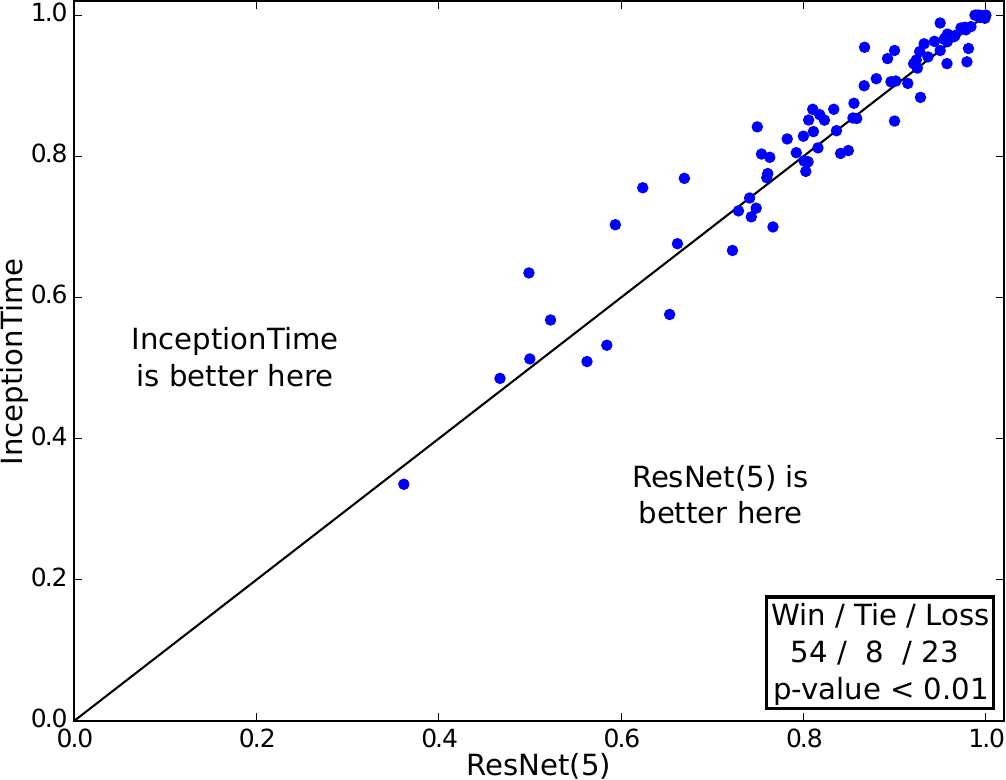}
    \caption{Plot showing how \ourmethod{} significantly outperforms ResNet(5).}
    \label{fig:plot-inception-vs-resnet}
\end{figure}

The pairwise accuracy plot in \figurename~\ref{fig:plot-inception-vs-resnet} compares \ourmethod{} to a model we call ResNet(5), which is an ensemble of 5 different ResNet networks~\citep{IsmailFawaz2019deep}. 
We found that \ourmethod{} showed a significant improvement over its neural network competitor, the previous best deep learning ensemble for TSC.
Specifically, our results show a Win/Tie/Loss of 54/8/23 in favor of \ourmethod{} against ResNet(5) with a $p$-value $<0.01$, suggesting the significant gain in performance is mainly due to improvements in our proposed Inception network architecture.
Additionally, in order to have a fair comparison between ResNet(5) and \ourmethod{}, we fixed the batch size of ResNet to 64 -- equal to the default value used for \ourmethod{}. 
This would further highlight that the improvement is mainly due to the architectural design of our proposed network, and not due to some other optimization hyperparameter such as the batch size.
Finally, we would like to note that when using the original batch size value proposed by~\cite{wang2017time} for ResNet, we observed similar results: \ourmethod{} was significantly better than the original ResNet(5) with a Win/Tie/Loss of 53/7/25.

\begin{figure}
    \centering
    \includegraphics[width=.7\linewidth]{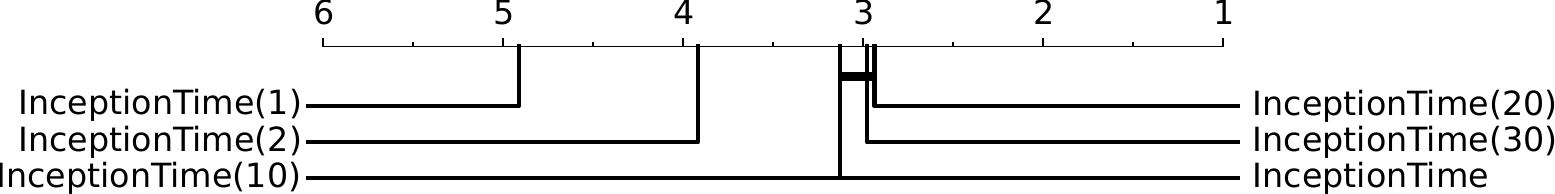}
    \caption{Critical difference diagram showing the effect of the number of individual classifiers in the \ourmethod{} ensemble.}
    \label{fig:cd-diagram-ensembles}
\end{figure}

In order to better understand the effect of the randomness on the accuracy of our neural networks, we present in \figurename~\ref{fig:cd-diagram-ensembles} the critical difference diagram of different \ourmethod{}($x$) ensembles with $x\in \{1,2,5,10,20,30\}$ denoting the number of individual networks in the ensemble.
Note that \ourmethod{}(1) is equivalent to a single Inception network and \ourmethod{} is equivalent to \ourmethod{}(5).
By observing \figurename~\ref{fig:cd-diagram-ensembles} we notice how there is no significant improvement when $x\ge5$, which is why we chose to use an ensemble of size 5, to minimize the classifiers' training time. 

\section{Architectural Hyperparameter study}\label{sec:hyp}

In this section, we will further investigate the hyperparameters of our deep learning architecture and the characteristics of the Inception module in order to provide insight for practitioners looking at optimizing neural networks for TSC.
First, we start by investigating the batch size hyperparameter, since this will greatly influence training time of all of our models.
Then we investigate the effectiveness of residual and bottleneck connections, both of which are present in \ourmethod{}.
After this we will experiment on model depth, filter length, and number of filters.
In all experiments the default values for \ourmethod{} are: batch size 64; bottleneck size 32; depth 6; filter length \{10,20,40\}; and, number of filters 32.
\revisionn{Finally, since the train/test split (provided in the archive) does not help in estimating the generalization ability of our approach, we have conducted a sensitivity analysis that evaluates the second best value for each of the network's hyperparameters (see subsection~\ref{sens}).}

\subsection{Batch size}
\begin{figure}
    \centering
    \includegraphics[width=.7\linewidth]{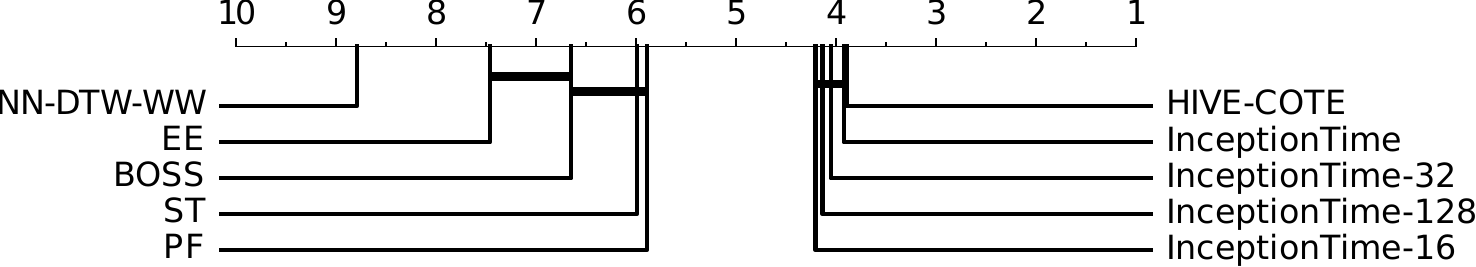}
    \caption{Critical difference diagram showing the effect of the batch size hyperparameter value over \ourmethod{}'s average rank.}
    \label{fig:cd-diagram-inception-batch-size}
\end{figure}

We started by investigating the batch size hyperparameter on the UCR archive, since this will greatly influence the training time of our models.
The critical difference diagram in \figurename~\ref{fig:cd-diagram-inception-batch-size} shows how the batch size affects the performance of \ourmethod{}. 
The horizontal thick line between the different models shows a non significant difference between them when evaluated on the 85 datasets, with a small superiority to \ourmethod{} (batch size equal to 64).
Finally, we should note that as we did not observe any significant impact on accuracy we did not study the effect of this hyperparameter on the simulated dataset and we chose to fix the batch size to 64 (similarly to \ourmethod{}) when experimenting on the simulated dataset below.

\subsection{Bottleneck and residual connections}

\begin{figure}
    \centering
    \includegraphics[width=0.6\linewidth]{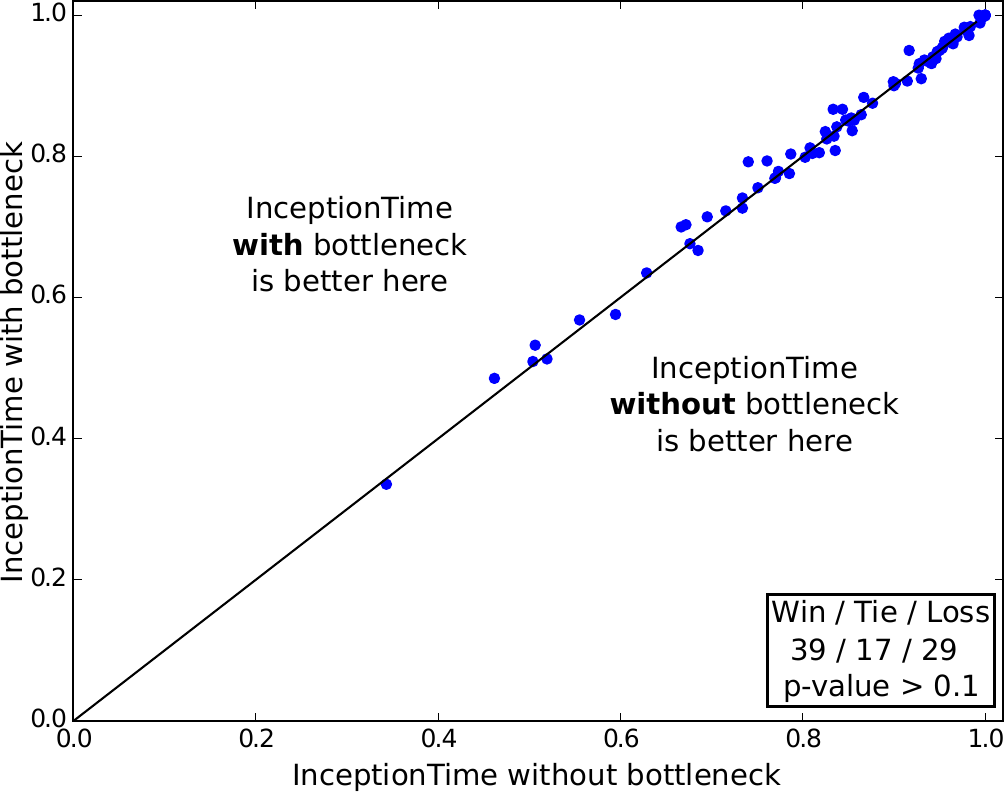}
    \caption{Accuracy plot for \ourmethod{} with/without the bottleneck layer.}
    \label{fig:plot-inception-with-without-bottleneck}
\end{figure}

In~\cite{ismailfawaz2018deep}, compared to other deep learning classifiers, ResNet achieved the best classification accuracy when evaluated on the 85 datasets and as a result we chose to look at the specific characteristic of this architecture --- its residual connections. 
Additionally, we tested one of the defining characteristics of Inception --- the bottleneck feature. 
For the simulated dataset, we did not observe any significant impact of these two connections, we therefore proceed with experimenting on the 85 datasets from the UCR archive.

\figurename~\ref{fig:plot-inception-with-without-bottleneck} shows the pairwise accuracy plot comparing \ourmethod{} with/without the bottleneck.
Similar to the experiments on the simulated dataset, we did not find any significant variation in accuracy when adding or removing the bottleneck layer.

\begin{figure}
    \centering
    \includegraphics[width=.7\linewidth]{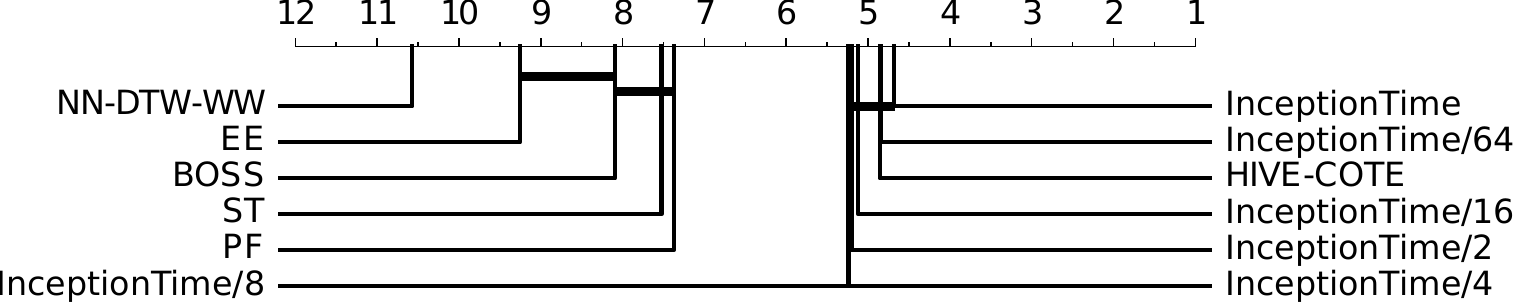}
    \caption{Critical difference diagram showing how the network's bottleneck size affects \ourmethod{}' average rank.}
    \label{fig:cd-diagram-inceptiontime-bottleneck-size-with-bake-off}
\end{figure}

In fact, using a Wilcoxon Signed-Rank test we found that  \ourmethod{} with the bottleneck layer is only slightly better than removing the bottleneck layer ($p$-value $>0.1$).
In terms of accuracy, these results all suggest not to use a bottleneck layer, however we should note that the major benefit of this layer is to significantly decrease the number of parameters in the network. 
In this case, \ourmethod{} with the bottleneck contains almost half the number of parameters to be learned, and given that it does not significantly decrease accuracy, we chose to retain its usage.
In a more general sense, these experiments suggest that choosing whether or not to use a bottleneck layer is actually a matter of finding a balance between a model's accuracy and its complexity.
The latter observation is evident in \figurename~\ref{fig:cd-diagram-inceptiontime-bottleneck-size-with-bake-off} where choosing smaller bottleneck size in order to reduce \ourmethod{}'s runtime will result in small yet insignificant decrease in accuracy.

\begin{figure}
    \centering
    \includegraphics[width=0.6\linewidth]{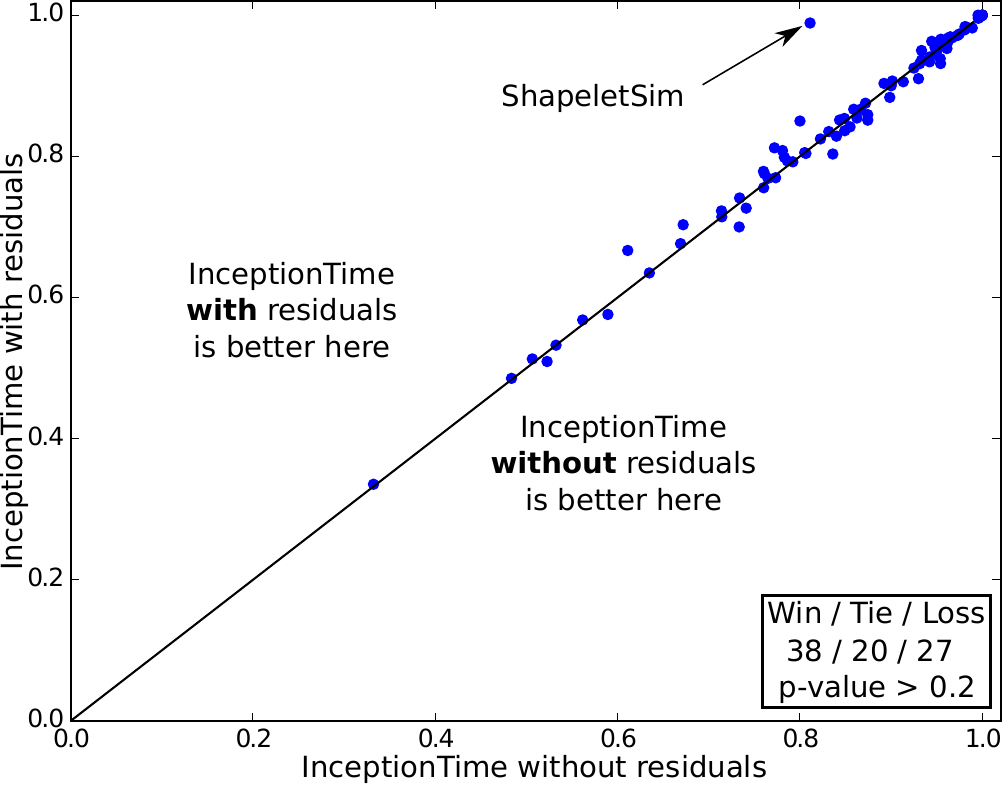}
    \caption{Accuracy plot for \ourmethod{} with/without the residual connections.}
    \label{fig:plot-inception-with-without-residual}
\end{figure}

To test the residual connections, we simply removed the residual connection from \ourmethod{}. 
Thus, without any shortcut connection, \ourmethod{} will simply become a deep convolutional neural network with stacked Inception modules.
\figurename~\ref{fig:plot-inception-with-without-residual} shows how the residual connections have a minimal effect on accuracy when evaluated over the whole 85 datasets in the UCR archive with a $p$-value $>0.2$.  

This result was unsurprising given that for computer vision tasks residual connections are known to improve the convergence rate of the network but not alter its test accuracy~\citep{szegedy2017inception}. 
However, for some datasets in the archive, the residual connections did not show any improvement nor deterioration of the network's convergence either. This could be linked to other factors that are specific to these data, such as the complexity of the dataset.

One example of interest that we noticed was a significant decrease in \ourmethod{}'s accuracy when removing the residual component for the ShapeletSim dataset. 
This is a synthetic dataset, designed specifically for shapelets discovery algorithms, with shapelets (discriminative subsequences) of different lengths~\citep{hills2014classification}.
Further investigations on this dataset indicated that \ourmethod{} without the residual connections suffered from a severe overfitting. 

While not the case here, some research has observed benefits of skip, dense or residual connections~\citep{huang2017densely}.
Given this, and the small amount of labeled data available in TSC compared to computer vision problems, we believe that each case should be independently studied whether to include residual connections.
\revisionn{The latter observation suggests that a large scale general purpose labeled dataset similar to ImageNet~\citep{russakovsky2015imagenet} is needed for TSC.}
Finally, we should note that the residual connection has a minimal impact on the network's complexity~\citep{szegedy2017inception}. 

\subsection{Depth}

Most of deep learning's success in image recognition tasks has been attributed to how `deep' the architectures are~\citep{lecun2015deep}. 
Consequently, we decided to further investigate how the number of layers affects a network's accuracy.
Unlike the previous hyperparameters, we present here the results on the simulated dataset. 
Apart from the depth parameter, we used the default values of \ourmethod{}.
For this dataset we fixed the number of training instances to 256 and the number of classes to 2 (see \figurename~\ref{fig:synthetic-dataset-example} for an example).
The only dataset parameter we varied was the length of the input time series.

\begin{figure}
    \centering
    \includegraphics[width=.6\linewidth]{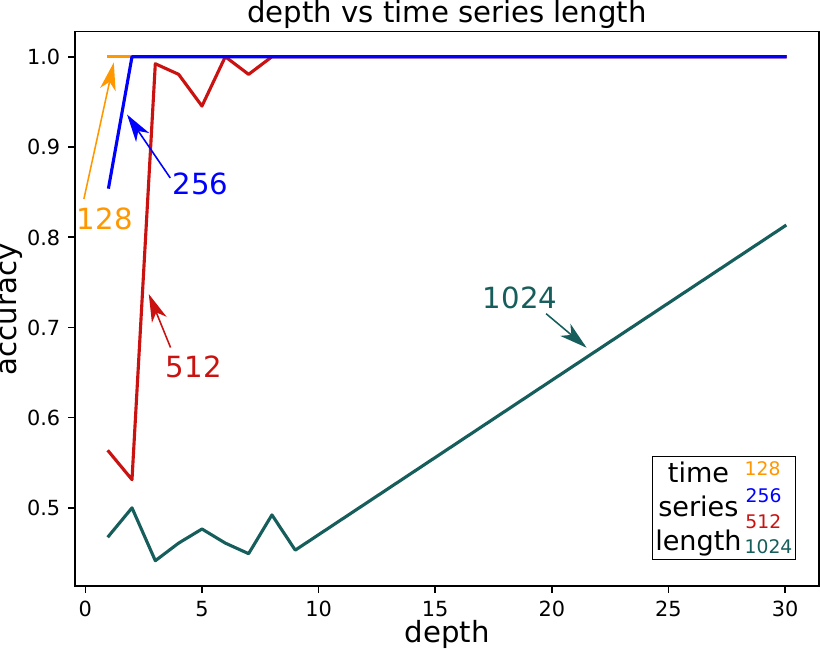}
    \caption{Inception network's accuracy over the simulated dataset, with respect to the network's depth as well as the length of the input time series.}
    \label{fig:depth-vs-length}
\end{figure}

\figurename~\ref{fig:depth-vs-length} illustrates how the model's accuracy varies with respect to the network's depth when classifying datasets of time series with different lengths. 
Our initial hypothesis was that as longer time series can potentially contain longer patterns and thus should require longer receptive fields in order for the network to separate the classes in the dataset. 
In terms of depth, this means that longer input time series will garner better results with deeper networks. 
And indeed, when observing \figurename~\ref{fig:depth-vs-length}, one can easily spot this trend: deeper networks deliver better results for longer time series.

\begin{figure}
    \centering
    \includegraphics[width=.7\linewidth]{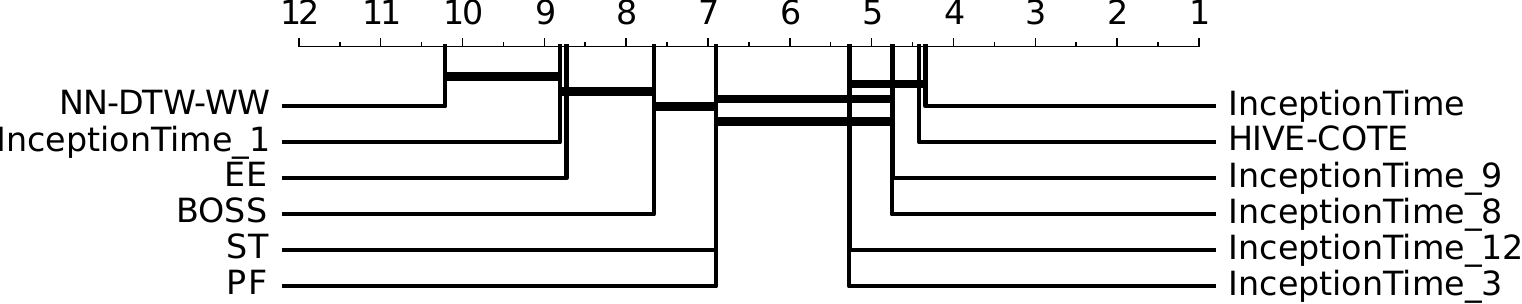}
    \caption{Critical difference diagram showing how the network's depth affects \ourmethod{}' average rank.}
    \label{fig:cd-diagram-inceptiontime-depth-with-bake-off}
\end{figure}

In order to further see how much effect the depth of a model has on real TSC datasets, we decided to implement deeper and shallower \ourmethod{} models, by varying the depth between 1 layer and 12 layers.
In fact, compared with the original architecture proposed by~\cite{wang2017time}, the deeper (shallower) version of \ourmethod{} will contain one additional (fewer) residual blocks each one comprised of three inception modules. 
By adding these layers, the deeper (shallower) \ourmethod{} model will contain roughly double (half) the number of parameters to be learned.
\figurename~\ref{fig:cd-diagram-inceptiontime-depth-with-bake-off} depicts the critical difference diagram comparing the deeper and shallower \ourmethod{} models to the original \ourmethod{}.

Unlike the experiments on the simulated dataset, we did not manage to improve the network's performance by simply increasing its depth. 
This may be due to many reasons, however it is likely due to the fact that deeper networks need more data to achieve high generalization capabilities~\citep{lecun2015deep}, and since the UCR archive does not contain datasets with a huge number of training instances, the deeper version of \ourmethod{} was overfitting the majority of the datasets and exhibited a small insignificant decrease in performance.
On the other hand, the shallower version of \ourmethod{} suffered from a significant decrease in accuracy (see \ourmethod{}\_3 and \ourmethod{}\_1 in \figurename~\ref{fig:cd-diagram-inceptiontime-depth-with-bake-off}). 
This suggests that a shallower architecture will contain a significantly smaller RF, thus achieving lower accuracy on the overall UCR archive.

From these experiments we can conclude that increasing the RF by adding more layers will not necessarily result in an improvement of the network's performance, particularly for datasets with a small training set.
However, one benefit that we have observed from increasing the network's depth, is to choose an RF that is long enough to achieve good results without suffering from overfitting.

We therefore proceed by experimenting with varying the RF by varying the filter length.

\subsection{Filter length}

\begin{figure}
    \centering
    \includegraphics[width=0.6\linewidth]{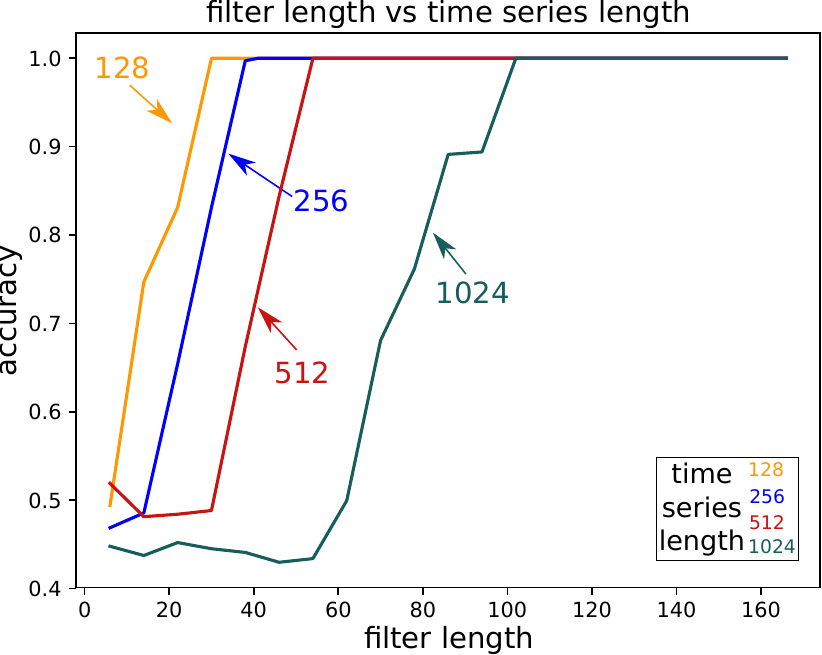}
    \caption{Inception network's accuracy over the simulated dataset, with respect to the filter length as well as the input time series length.}
    \label{fig:filter-vs-length}
\end{figure}

In order to test the effect of the filter length, we start by analyzing how the length of a time series influences the accuracy of the model when tuning this hyperparameter. 
In these experiments we fixed the number of training time series to 256 and the number of classes to 2. \figurename~\ref{fig:filter-vs-length} illustrates the results of this experiment.

We can easily see that as the length of the time series increases, a longer filter is required to produce accurate results. 
This is explained by the fact that longer kernels are able to capture longer patterns, with higher probability, than shorter ones can.
Thus, we can safely say that longer kernels almost always improve accuracy.

\begin{figure}
    \centering
    \includegraphics[width=0.6\linewidth]{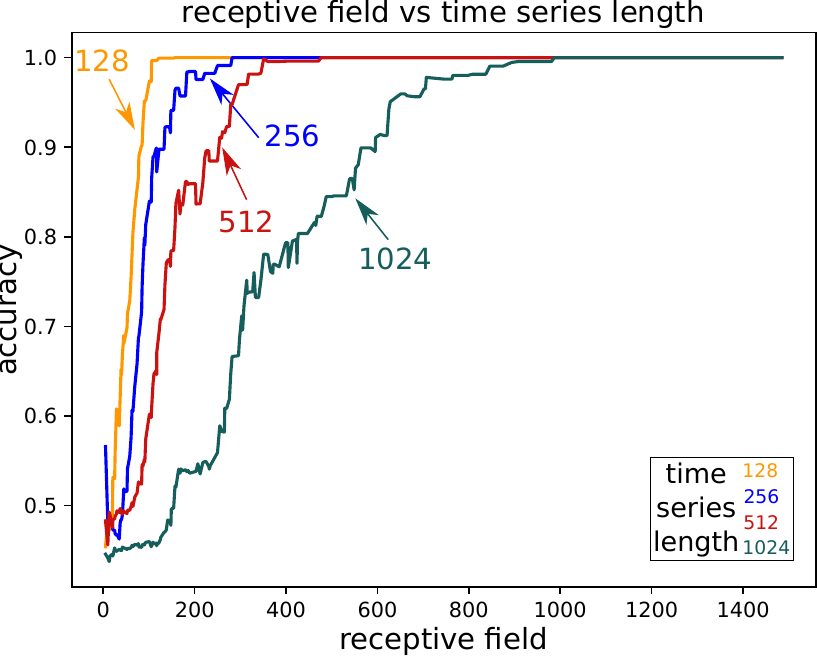}
    \caption{Inception network's accuracy over the simulated dataset, with respect to the receptive field as well as the input time series length.}
    \label{fig:plot-receptive-field}
\end{figure}

In addition to having visualized the accuracy as a function of both depth (\figurename~\ref{fig:depth-vs-length}) and filter length (\figurename~\ref{fig:filter-vs-length}), we proceed by plotting the accuracy as function of the RF for the simulated time series dataset with various lengths.
By observing \figurename~\ref{fig:plot-receptive-field} we can confirm the previous observations that longer patterns require longer RFs, with length clearly having a higher impact on accuracy compared to the network's depth.  
Moreover, by using a large enough RF to cover the whole input time series, the usage of a GAP layer won't affect \ourmethod{}'s ability to discriminate between the two patterns, because performing a GAP does not affect the model's RF.

\begin{figure}
    \centering
    \includegraphics[width=.7\linewidth]{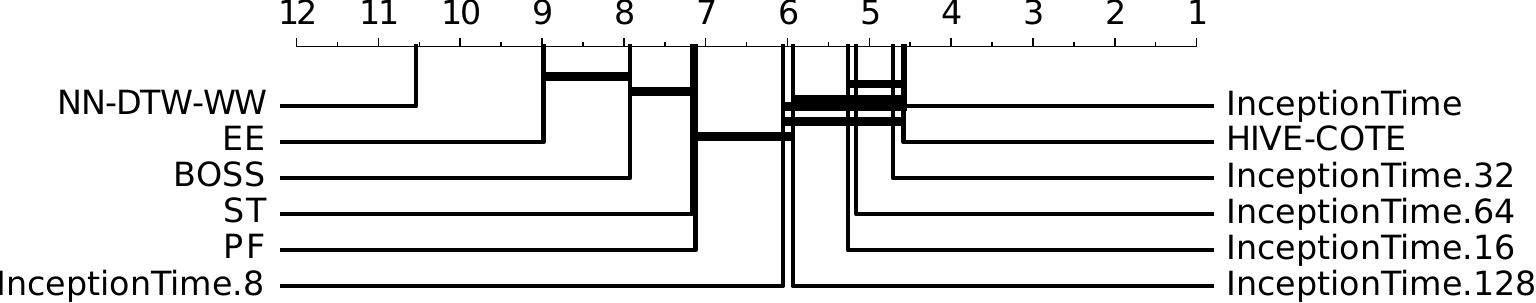}
    \caption{Critical difference diagram showing the effect of the filter length hyperparameter value over \ourmethod{}' average rank.}
    \label{fig:cd-diagram-inceptiontime-filtersize-with-bake-off}
\end{figure}

\begin{table}
\centering
\setlength\tabcolsep{4pt}
{ \scriptsize
\begin{tabularx}{0.60\textwidth}{cccccccccc}
\hline \\
\small Length     & \small \ourmethod{}.8  & \small \ourmethod{}.64  & \small \ourmethod{}  \\
\midrule \\ 
\small $<$81   & \textbf{1.71} & 2.21 & 1.79   \\
\small 81-250   & 1.89 & 2.11 & \textbf{1.42}   \\
\small 251-450  & 2.45 & \textbf{1.32} & 1.86   \\
\small 451-700  & 2.08 & 1.85 & \textbf{1.62} \\
\small 701-1000 & \textbf{1.50}  & 2.60  & 1.80  \\
\small $>$1000 & 2.14 & 2.00 & \textbf{1.71} \\
\hline
\end{tabularx}
}
\caption{Filter length variants of \ourmethod{} with their corresponding average ranks grouped by the datasets' length.
Bold indicates the best model.}\label{tab-perf-lengths}
\end{table}

There is a downside to longer filters however, in the potential for overfitting small datasets, as longer filters significantly increase the number of parameters in the network.
To answer this question, we again extend our experiments to the real data from the UCR archive, allowing us to verify whether long kernels tend to overfit the datasets when a limited amount of training data is available.
Therefore, we decided to train and evaluate \ourmethod{} versions containing both long and short filters on the UCR archive.
Where the original \ourmethod{} contained filters of length \{10,20,\textbf{40}\}, the five models we are testing here contain filters of length \{2,4,\textbf{8}\}; \{4,8,\textbf{16}\}; \{8,16,\textbf{32}\}; \{16,32,\textbf{64}\}; \{32,64,\textbf{128}\}.
\figurename~\ref{fig:cd-diagram-inceptiontime-filtersize-with-bake-off} illustrates a critical difference diagram showing how \ourmethod{} with longer filters will slightly decrease the network's performance in terms of accurately classifying the time series datasets.
We also investigate the relationship between the length of the time series and the length of the network's filter. 
\tablename~\ref{tab-perf-lengths} depicts the average rank of each variant of \ourmethod{} over the UCR archive grouped by the datasets' lengths (with about 15 datasets in each group).
Similarly to \figurename~\ref{fig:cd-diagram-inceptiontime-filtersize-with-bake-off}, we observe that almost for all time series length, \ourmethod{} with its default filter length (32) achieves the best or the second best overall accuracy.
We can therefore summarize that the results from the simulated dataset do generalize (to some extent) to real datasets: longer filters will improve the model's performance as long as there is enough training data to mitigate the overfitting phenomena.

In summary, we can confidently state that increasing the receptive field of a model by adopting longer filters will help the network in learning longer patterns present in longer time series. However there is an accompanying disclaimer that it may negatively impact the accuracy for some datasets due to overfitting.

\subsection{Number of filters}

\begin{figure}
    \centering
    \includegraphics[width=0.6\linewidth]{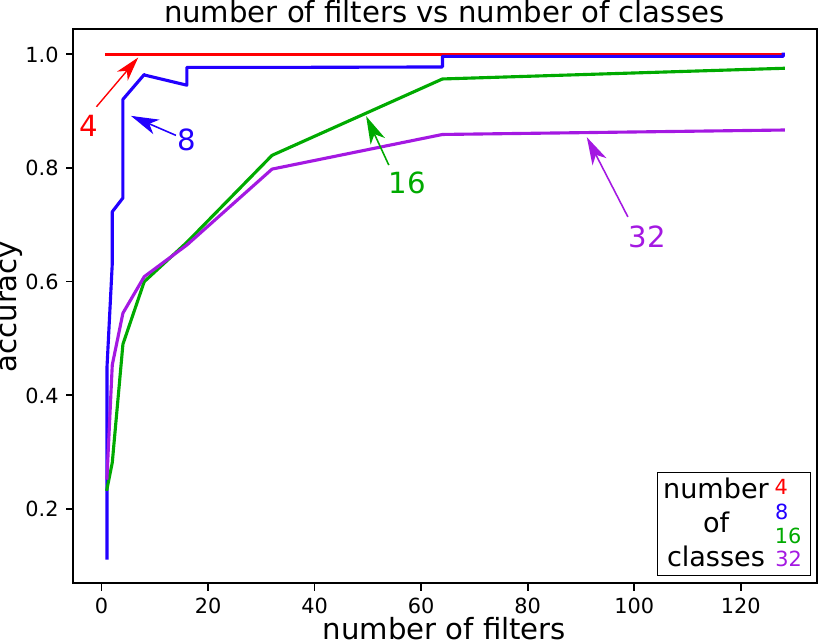}
    \caption{Inception network's accuracy over the simulated dataset, with respect to the number of filters as well as the number of classes.
    }
    \label{fig:nb-filters-vs-nb-classes}
\end{figure}

To provide some directions on how the number of filters affects the performance of the network, we experimented with varying this hyperparameter with respect to the number of classes in the dataset. 
To generate new classes in the simulated data, we varied the position and length of the patterns; for example, to create data with three classes, we inject patterns of the same length at three different positions.
For this series of experiments, we fixed the length of the time series to 256 and the number of training examples to 256.
 
\figurename~\ref{fig:nb-filters-vs-nb-classes} depicts the network's accuracy with respect to the number of filters for datasets with a differing number of classes.
Our prior intuition was that the more classes, or variability, present in the training set, the more features are required to be extracted in order to discriminate the different classes, and this will necessitate a greater number of filters. 
This is confirmed by the trend displayed in \figurename~\ref{fig:nb-filters-vs-nb-classes}, where the datasets with more classes require more filters to be learned in order to be able to accurately classify the input time series.

\begin{figure}
    \centering
    \includegraphics[width=.7\linewidth]{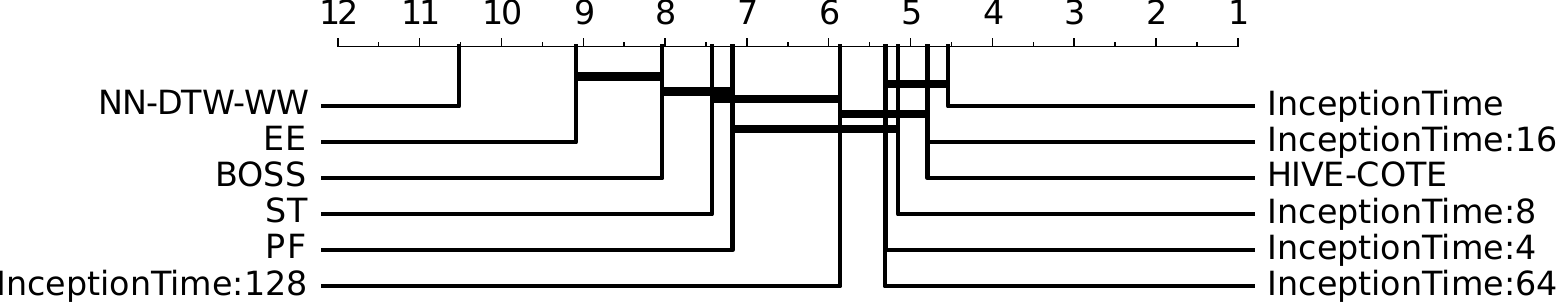}
    \caption{Critical difference diagram showing how the network's width affects \ourmethod{}' average rank.}
    \label{fig:cd-diagram-inceptiontime-nb-filters-with-bake-off}
\end{figure}

After observing on the synthetic dataset that the number of filters significantly affects the performance of the network, we asked ourselves if the current implementation of \ourmethod{} could benefit/lose from a naive increase/decrease in the number of filters per layer.
Our proposed \ourmethod{} model contains 32 filters per Inception module's component, while for these experiments we tested six ensembles, by varying the hyperparameter with a power of two.
\figurename~\ref{fig:cd-diagram-inceptiontime-nb-filters-with-bake-off} illustrates a critical difference diagram showing how increasing the number of filters per layer significantly deteriorated the accuracy of the network, whereas decreasing the number of filters did not significantly affect the accuracy. 
It appears that our \ourmethod{} model contains enough filters to separate the classes of the 85 UCR datasets, of which some have up to 60 classes (ShapesAll dataset).

Increasing the number of filters also has another side effect: it causes an explosion in the number of parameters in the network. 
The wider \ourmethod{} contains four times the number of parameters than the original implementation.
We therefore conclude that naively increasing the number of filters is actually detrimental, as it will drastically increase the network's complexity and eventually cause overfitting. 

\subsection{Sensitivity analysis}\label{sens}
Working with open benchmarks such as the UCR archive has pushed the community towards publishing high quality TSC algorithms. 
The UCR archive provides a train/test split for the data, which has allowed researchers to directly benchmark their works with the ones of others, as well as providing splits that were potentially more challenging and realistic than assuming that both train and test data were sampled from the same population. 
Having the train/test split available has however also led to the potential issue that the techniques designed on this benchmark archive might overfit it. 
This is especially true of deep learning classifiers that contain dozens of optimization and architectural hyperparameters~\citep{ismailfawaz2018deep}. 

\begin{figure}
    \centering
    \includegraphics[width=.7\linewidth]{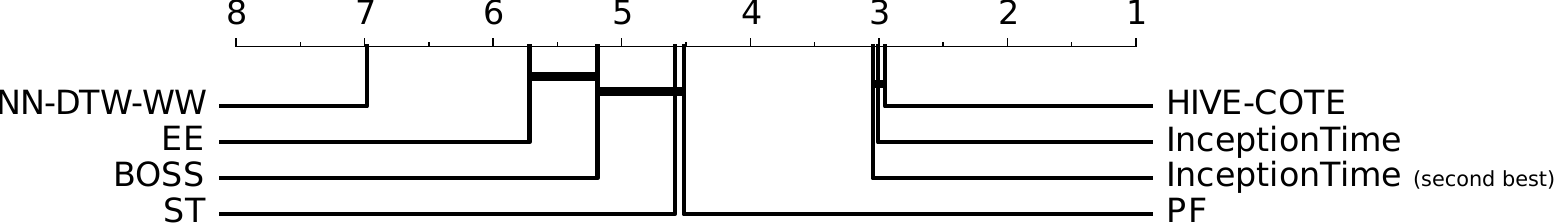}
    \caption{Critical difference diagram showing how choosing the second best hyperparameters affects \ourmethod{}'s average rank.}
    \label{fig:cd-diagram-inceptiontime-second-best}
\end{figure}

In an effort to give an idea of the sensitivity of \ourmethod{} to changes in its parameters, we have evaluated the performance of having chosen the second-best value of each of its parameters, that is the second-best value for the depth of the network (i.e. a value of 9 instead of the best value of 6), for its width (i.e. 16 instead of 32), for the length of the convolutions (final value of 32 instead of 40), for the batch size (i.e. 32 instead of 64), and for the bottleneck size (i.e. 64 instead of the default one 32). 
This gave us a new architecture ---~$\textsf{\ourmethod{}}_\textsf{\scriptsize{(second best)}}$~--- which we then compared with \ourmethod{} and also other algorithms. 
\figurename~\ref{fig:cd-diagram-inceptiontime-second-best} depicts the average rank of current state-of-the-art TSC algorithms with both \ourmethod{}'s default and second best hyperparameters added to the mix. 
We can clearly see that the effect is minimal: the ranking is a tiny bit lower but they are all well within the critical difference with HIVE-COTE (a post-hoc statistical test fails to reject the null hypothesis (p-value $\approx 0.71$) making the difference between the default and second best hyperparameters non significant). 

\section{Conclusion}\label{sec:conclusion}

Deep learning for time series classification still lags behind neural networks for image recognition in terms of experimental studies and architectural designs. 
In this paper, we fill this gap by introducing \ourmethod{}, inspired by the recent success of Inception-based networks for various computer vision tasks. 
We ensemble these networks to produce new state-of-the-art results for TSC on the 85 datasets of the UCR archive. 
Our approach is highly scalable, two orders of magnitude faster than current state-of-the-art models such as HIVE-COTE. 
The magnitude of this speed up is consistent across both Big Data TSC repositories as well as longer time series with high sampling rate. 
We further 
investigate the effects on overall accuracy of various hyperparameters of the CNN architecture. 
For these, we go far beyond the standard practices for image data, and design networks with long filters. 
We look at these by using a simulated dataset and frame our investigation in terms of the definition of the receptive field for a CNN for TSC.
\revisionn{Finally, although \ourmethod{} can be extended straightforwardly to multivariate data~\citep{ismailfawaz2018deep}, we would like to further explore applying to multivariate TSC the many architectural advancements in deep neural networks that are being published each year for computer vision tasks.}

\begin{acknowledgements}
The authors would like to thank the creators and providers of the datasets.
The authors would also like to thank NVIDIA Corporation for the GPU Grant and the M\'esocentre of Strasbourg for providing access to the cluster.
This work was supported by the ANR TIMES project (grant ANR-17-CE23-0015) of the French Agence Nationale de la Recherche.
François Petitjean is the recipient of an Australian Research Council Discovery Early Career Award
(project number DE170100037) funded by the Australian Government. This material is based upon work supported by the Air Force Office of Scientific Research, Asian Office of Aerospace Research and Development (AOARD) under award number FA2386-18-1-4030.
\end{acknowledgements}

\bibliographystyle{spbasic}      
\bibliography{biblio}

\end{document}